\documentclass[10pt,journal,compsoc]{IEEEtran}

\ifCLASSOPTIONcompsoc
  \usepackage[nocompress]{cite}
\else
  \usepackage{cite}
\fi
\usepackage{amsmath,amsfonts}
\usepackage{algorithmic}
\usepackage{algorithm}
\usepackage{array}
\usepackage[caption=false,font=normalsize,labelfont=sf,textfont=sf]{subfig}
\usepackage{textcomp}
\usepackage{stfloats}
\usepackage{url}
\usepackage{verbatim}
\usepackage{graphicx}
\usepackage{cite}
\usepackage{booktabs}
\usepackage{multirow}
\usepackage{todonotes}

\ifCLASSINFOpdf
\else
\fi
\begin{document}
%
\title{
Learning Group Interactions and Semantic Intentions for Multi-Object Trajectory Prediction}

%
%
%
%

\author{Mengshi Qi,~\IEEEmembership{Member,~IEEE,}
        Yuxin Yang,
        Huadong Ma,~\IEEEmembership{Fellow,~IEEE} 
\thanks{This work is partly supported by the Funds for the NSFC Project under Grant 62202063, Beijing Natural Science Foundation (L243027), the Innovation Research Group Project of the NSFC under Grant 61921003. (\emph{Corresponding author: Mengshi Qi~(email:~qms@bupt.edu.cn)})}
\thanks{M. Qi, Y. Yang, and H. Ma are with State Key Laboratory of Networking and Switching Technology, Beijing University of Posts and Telecommunications, China.}
}

\IEEEtitleabstractindextext{

\begin{abstract}
Effective modeling of group interactions and dynamic semantic intentions is crucial for forecasting behaviors like trajectories or movements. In complex scenarios like sports, agents' trajectories are influenced by group interactions and intentions, including team strategies and opponent actions. To this end, we propose a novel diffusion-based trajectory prediction framework that integrates group-level interactions into a conditional diffusion model, enabling the generation of diverse trajectories aligned with specific group activity. To capture dynamic semantic intentions, we frame group interaction prediction as a cooperative game, using Banzhaf interaction to model cooperation trends. We then fuse semantic intentions with enhanced agent embeddings, which are refined through both global and local aggregation. Furthermore, we expand the NBA SportVU dataset by adding human annotations of team-level tactics for trajectory and tactic prediction tasks. Extensive experiments on three widely-adopted datasets demonstrate that our model outperforms state-of-the-art methods. Our source code and data are available at https://github.com/aurora-xin/Group2Int-trajectory.
\end{abstract}

}

\maketitle

\IEEEdisplaynontitleabstractindextext

%
\IEEEpeerreviewmaketitle

\ifCLASSOPTIONcompsoc

\IEEEraisesectionheading{\section{Introduction}\label{sec:introduction}}
\else
\section{Introduction}
\label{sec:introduction}
\fi


\IEEEPARstart{M}{ulti-object} trajectory prediction is a task that aims to forecast multi-modal future movements in the scene based on their observed history and contextual information. It is essential to obtain accurate trajectory prediction for decision-making in various real-world applications, including sports analysis systems~\cite{felsen2018will, qi2020imitative,qi2018stagnet,xu2022groupnet}, autonomous driving systems~\cite{zhou2023query,alahi2016social,huang2019stgat,rasouli2021bifold}, social robots~\cite{cheng2020towards}, etc.

\begin{figure}[h]
    \centering
    \includegraphics[width=0.9\linewidth]{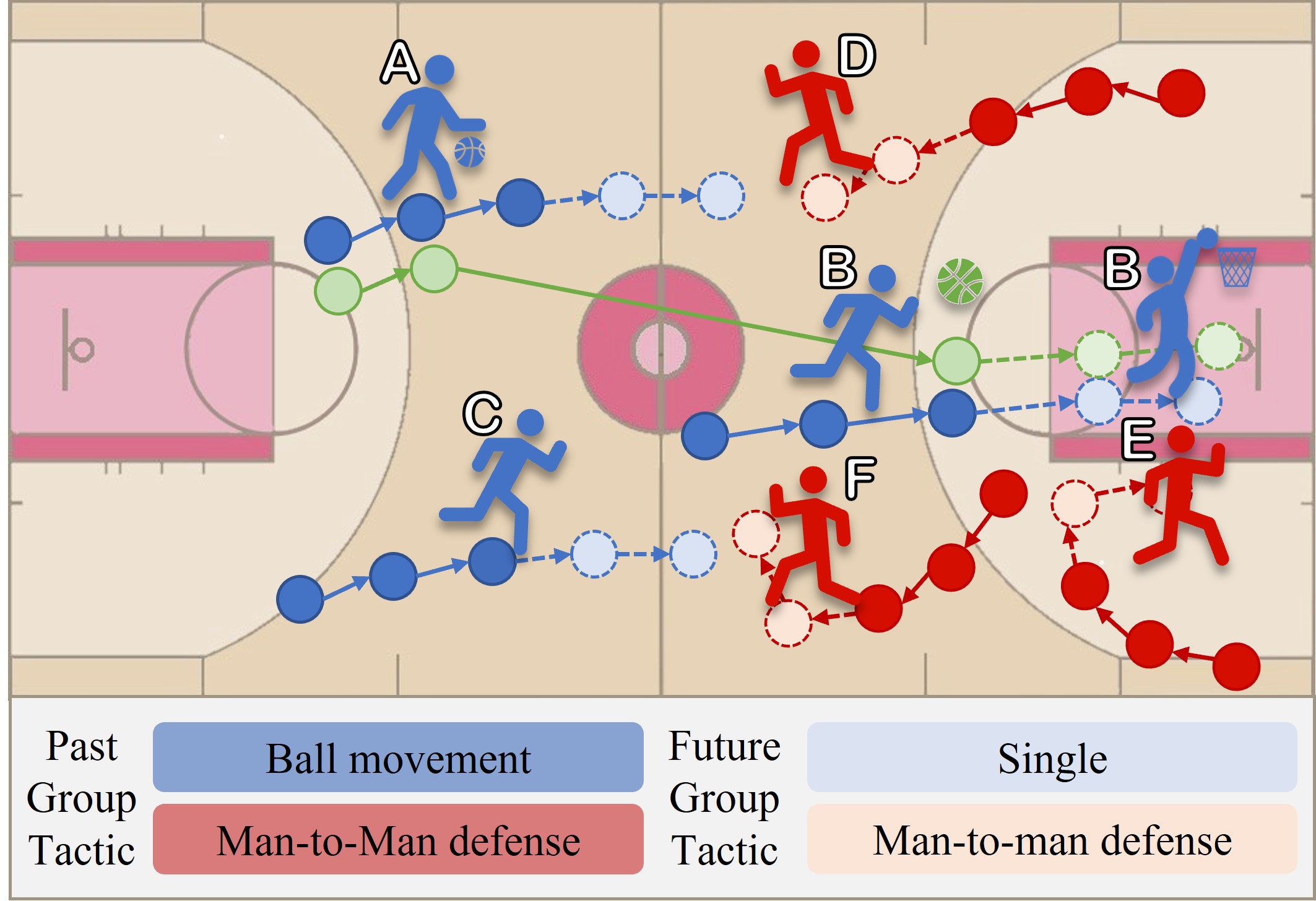}
     \caption{Illustration of multi-agent trajectory prediction in a 3 vs. 3 basketball game. Agent trajectories are shown as circles, blue for one team, red for the opposing team, and green for the basketball. Solid circles are observed trajectories, while dashed ones are predicted positions. The blue team begins with a "Ball Movement" tactic, switching to "Single", while the red team uses "Man-to-Man Defense". Given the observed 2D trajectories and tactics, our goal is to (a) predict future trajectories and (b) forecast the tactics each team will adopt in the next frames.}
    \label{fig:intro}
\end{figure}

Especially, trajectory prediction in sports competitions~\cite{felsen2018will, qi2020imitative,qi2018stagnet,xu2022groupnet} presents unique challenges and opportunities due to the complex group interactions among objects and the rapidly changing semantic intention demands of the game. Incorporating expert knowledge, such as tactics, can enhance both prediction accuracy and interpretability, thereby providing invaluable support for tactical analysis and decision-making. We present a case study of a 3 vs. 3 basketball game in Fig.~\ref{fig:intro}. 

From the figure, we can see the blue team, on offense and advancing to the right, initially employs a ''Ball Movement`` tactic for prior three frames, in which player A pass to player B near the basket to prevent interception by player D. Then the tactic transitions to a "Single" play over the next two frames, as player B has an opportunity for a quick shot. While the red team implements a "Man-to-Man Defense" with each defender closely following their assigned opponent, where player E try to obstruct the movement of player B. Recognizing the blue team's shift from "Ball Movement" to "Single" indicates the ball-handler's intent to drive or shoot, while understanding the red team's defense suggests that defensive trajectories will mirror the movements of offensive players, demonstrating learning strategy transitions informs movement trajectories and underlying motivations.

As deep learning advances rapidly in recent years, there have been generative model based attempts to represent multi-modality of human motion by latent variable, such as generative adversarial networks (GANs)\cite{gupta2018socialgan,qi2020imitative} and conditional variational autoencoders (CVAEs)\cite{salzmann2020trajectron++,mangalam2020not,xu2022groupnet}. However, due to limitations like unstable training process of GANs and unnatural trajectories generation of CVAE~\cite{gu2022stochastic}, diffusion-based methods have been utilized for this specific task. Recent advancements in diffusion-based trajectory prediction leverage observed trajectories as conditions~\cite{gu2022stochastic}, enhancing inference speed~\cite{mao2023leapfrog} with a trainable initializer and human motion representation~\cite{bae2024singulartrajectory} in a singular space.
Meanwhile, existing methods have proposed various solutions that focus on modeling social interactions through pooling mechanisms~\cite{alahi2016social} and graph-based representations of spatial and temporal relationships~\cite{huang2019stgat,mohamed2020social,xu2022groupnet}.

However, these works have limitations in analyzing group-level interactions and exploring variations in semantic intention. Existing methods tend to focus on agent-level interactions but ignore higher-level group interactions and implicit dynamics of semantic intention changes, which are crucial for guiding agents to adopt different trajectories, especially in complex competitive sports settings. In sports, group interactions are diverse, including intra-team cooperation, responses to opponents, and mutual influences among team members. Such interactions often exhibit a game-theoretic nature, where each player’s intentions are shaped not only by team strategies but also by the actions and intentions of opponents. This complexity impacts the overall distribution of player trajectories. Thus, more comprehensive incorporation of intention variation and group-level interactions in trajectory prediction models can enable more accurate and realistic simulations of sports dynamics, enhancing predictive reliability. In conclusion, two major challenges have not been fully tackled in multi-agent trajectory forecasting: 1) how to model the complex group interactions among multiple objects; 2) how to capture semantic intention in dynamic scenarios to guide trajectory prediction.

In order to address aforementioned challenges, we propose a novel end-to-end trajectory prediction framework that leverages group interactions and semantic intentions for trajectory prediction. The framework incorporates expert knowledge for modeling group interactions and employs game theory to capture the dynamic semantic intentions among agents and groups, especially in sports games. Specifically, we utilize an interaction encoder to process observed trajectories with group-level tactics as condition for diffusion model in the classifier-free guidance~\cite{DBLP:journals/corr/cfg} manner, learning explicit group interactions for trajectory prediction. Moreover, we capture both global and local information through attention mechanisms to obtain enhanced agent tokens. More importantly, we model group interaction prediction as a cooperative game, where agents in the scene and the potential tactics each group adopts are considered as players according to the game theory. We utilize Banzhaf Interaction~\cite{jin2023video} to represent the cooperation trends among any coalition of players, reflecting the semantic intentions across multiple levels.
Finally, we fuse the enhanced agent tokens with the learned semantic intention assignments for each agent, derived from the learned similarity, and input to the prediction head for each group to obtain group interaction probabilities.

The main contributions can be summarized as follows:
\begin{enumerate}
    \item {We propose a novel diffusion-based trajectory prediction method that incorporates group-level interactions as expert knowledge, providing an additional condition to generate more realistic and interactive trajectories.}
    \item {We present an innovative semantic intentions prediction method that leverages the game theory to capture multi-level dynamic activity changes among various agents and groups, thereby enhancing the model's ability in reasoning.}
    \item {We introduce a new benchmark by expanding the NBA SportVU dataset with human annotations of team-level dynamic tactics for trajectroy and tactic prediction tasks. Extensive experiments on three widely-adopted datasets demonstrate our model’s state-of-the-art performance even compared to the Large Language Models~(LLMs).}
\end{enumerate}

\section{Related Works}
\subsection{Trajectory prediction}
Early trajectory prediction research primarily relies on deterministic models such as Markov processes~\cite{kumar2016ask}, social forces~\cite{helbing1995social} and recurrent neural networks~\cite{vemula2018social}, which incorporate social interactions among agents and highlight the role of generative models. 
As deep learning advances~\cite{qi2020stc,yun2024weakly,qi2019attentive}, stochastic prediction, which considers all potential future trajectories, such as an agent’s decisions at a crossroads, has emerged as the prevailing approach in the field.
These works combine probabilistic inferences with the bivariate Gaussian distribution~\cite{alahi2016social,li2020evolvegraph,mohamed2020social,mohamed2022social,shi2021sgcn,xu2022adaptive}, Generative Adversarial Network (GAN)~\cite{gupta2018socialgan,sadeghian2019sophie,dendorfer2021mg,sun2023stimulus}, Conditional Variational AutoEncoder (CVAE)~\cite{chen2021personalized,ivanovic2019trajectron,lee2022muse,lee2017desire,mangalam2020not} and Diffusion model~\cite{gu2022stochastic,mao2023leapfrog,rempe2023trace,jiang2023motiondiffuser,bae2024singulartrajectory,yang2024uncovering} for multi-modal trajectory generation.
For example, Social-LSTM~\cite{alahi2016social} introduces a social pooling layer to capture agent interactions, extended by Social-GAN~\cite{gupta2018socialgan}. Additionally, attention-based methods are essential to capture critical interactions in crowded environments~\cite{huang2019stgat}. Other works integrate scene understanding to extract global information, such as SS-LSTM~\cite{xue2018ss}. Trajectron++~\cite{salzmann2020trajectron++} establishes the connections between scene information and agent motion using graph structures, and MANTRA~\cite{marchetti2020mantra} combines memory mechanisms with scene images. 
MID~\cite{gu2022stochastic} is noteworthy as it introduced diffusion models to predict trajectories by modeling the process of human motion variation from indeterminate to determinate, and laying the foundation of Transformer diffusion for trajectory prediction that a serious of later works~\cite{mao2023leapfrog, bae2024singulartrajectory} adopt. Specifically, LED~\cite{mao2023leapfrog} proposes a leapfrog initializer to accelerate the trajectory prediction. 
TRACE~\cite{rempe2023trace} incorporates controllable guidance factors such as goals, avoidance, and social groups into the diffusion model, leading to the generation of trajectory data that rivals real-world scenarios while maintaining rationality, SingularTrajectory~\cite{bae2024singulartrajectory} unifies human motion representations in a singular space to help generate trajectories.
Moreover, existing works in the sports domain~\cite{qi2020imitative, kim2023ball, hauri2021multi,qi2019sports,yun2025semi,qi2021semantics} also focus on trajectory forecasting and imputations, analyzing and modeling sports activities and agent behaviors. However, most of these works emphasis on interactions among agents, but ignore the dynamic group activity and semantic intention changes. Different from these previous works, we model the group interactions in a classifier-free guidance manner for diffusion model in sports scenarios, using the detected tactic strategy as an additional group-level condition to enhance trajectory prediction performance.

\subsection{Diffusion models}
Diffusion models~\cite{ho2020denoising} have demonstrated promising results across various generative applications, including image generation~\cite{ho2020denoising,song2020denoising,DBLP:journals/corr/cfg} and natural language generation~\cite{austin2021structured}.
These models are a class of neural generative models that leverage stochastic diffusion processes based on thermodynamic principles. They iteratively introduce noise to data samples and train a neural network to reverse this process by removing the noise. Diffusion models have also shown great potential for generation in trajectory prediction~\cite{gu2022stochastic,mao2023leapfrog,bae2024singulartrajectory}. Specifically, MID~\cite{gu2022stochastic} is the first to utilize diffusion models for trajectory prediction, effectively modeling the indeterminacy variation process. LED~\cite{mao2023leapfrog} introduces a trainable initializer to establish an expressive distribution, thereby reducing the number of denoising steps required to accelerate inference. SingularTrajectory~\cite{bae2024singulartrajectory} unifies human motion representations in a singular space and employs an adaptive anchor within a diffusion model framework.
A seminal work that performs text-guided generation with diffusion is classifier-guidance diffusion~\cite{DBLP:journals/corr/cfg}. Classifier-free guidance (CFG)~\cite{DBLP:journals/corr/cfg} forms the key basis of modern text-guided generation with diffusion models~\cite{dhariwal2021diffusion, rombach2022high}.
However, it requires an accurate estimation of guidance gradient based on the classifier.
Thus, classifier-free guidance offers a significant advantage over classifier-guided models~\cite{dhariwal2021diffusion}. Then, a branch of this kind of conditional diffusion method achieves state-of-the-art performance in a variety of tasks~\cite{ajay2022conditional}. In our work, we adopt the classifier-free approach to augment the trajectory generation process with a guidance signal, which amplifies the features of specific hidden constraints and optimal group interactions.

\subsection{Cooperative Game Theory}

Game theory is widely applied in economic theory~\cite{rabin1992incorporating,kreps1990game}, politics~\cite{brams2011game}, but its exploration in computer vision domain, particularly human trajectory prediction, is relatively limited. The cooperative game theory involves a set of players and a characteristic function~\cite{osborne1994course}, which maps each possible subset of players (also known as a coalition) to a real number representing the total payoff achieved by those players working together toward a shared goal~\cite{jin2023video,ma2017forecasting,li2022fine}.
The core of cooperative game theory is to fairly and reasonably allocate different payoffs to individual players based on their contributions. Researchers introduce various value concepts, such as the Shapley value~\cite{shapley:book1952,WINTER20022025} and Banzhaf value~\cite{Banzhaf_1965_5380,lehrer1988axiomatization}, to measure the average added value that each player brings to different coalitions. Some prior works utilize the game theory to improve model interpretability~\cite{yang2022explaining}, while some works incorporate the game theory for the vision-language task~\cite{jin2023video,li2022fine}. Recently, GameFormer~\cite{huang2023gameformer} uses the hierarchical game theory to decode future trajectories for vehicle prediction and planning. In our work, we exploit the game theory to model dynamic semantic intentions in sports games, of which we incorporate Banzhaf Interaction~\cite{jin2023video} to obtain multi-level interactions, thereby improving the accuracy, reliability and interpretability of human trajectory prediction.

\section{Preliminary}
\subsection{Diffusion models}
The diffusion model converts a noisy distribution represented by the noise vector $\mathbf{y}_{T}$, into the desired clean data $\mathbf{y}_{0}$ over \(T\) steps, utilizing intermediate latent variables $\{\mathbf{y}_{t} \mid t \in [1, \ldots, T]\}$ that are involved in both the diffusion and denoising phases. During the diffusion phase, noise is progressively introduced into the data in incremental steps,  transforming the distribution $q(\mathbf{y}_{0})$ into the standard normal distribution $q(\mathbf{y}_{T})$, through a Markov chain as follows:
\begin{equation}
q(\mathbf{y}_{1:T} \mid \mathbf{y}_{0}) := \prod_{t=1}^{T} q(\mathbf{y}_{t} \mid \mathbf{y}_{t-1}),
\end{equation}
\begin{equation}
q(\mathbf{y}_{t} \mid \mathbf{y}_{t-1}) := \mathcal{N}(\mathbf{y}_{t}; \sqrt{1-\beta_t}\mathbf{y}_{t-1}, \beta_t \mathbf{I}),
\end{equation}
where $\beta_t$ is a constant that controls the variance schedule for noise injection. In the denoising phase, $\mathbf{y}_{t}$ is used to reconstruct $\mathbf{y}_{0}$ via a learnable model as:
\begin{equation}
p_\theta(\mathbf{y}_{0:T}) := p(\mathbf{y}_{T}) \prod_{t=1}^{T} p_\theta(\mathbf{y}_{t-1} \mid \mathbf{y}_{t}),
\end{equation}
\begin{equation}
p_\theta(\mathbf{y}_{t-1} \mid \mathbf{y}_{t}) := \mathcal{N}(\mathbf{y}_{t-1}; f_{\epsilon}(\mathbf{y}_{t}, t), \beta_t \mathbf{I}),
\end{equation}
where $\mathbf{y}_{T} \sim \mathcal{N}(0, \mathbf{I})$ denotes the initial noise sampled from the Gaussian distribution $p(\mathbf{y}_{T})$, $\theta$ represents the learnable parameters of the diffusion model and \(f_{\epsilon}\) is the learnable denoising model. The goal is to train the neural network to enable the denoising process to accurately approximate the true data distribution, typically by maximizing the evidence lower bound (ELBO), ensuring that the samples generated by the diffusion model closely resemble real data.

Furthermore, classifier-free guidance is well known to yield significantly improved samples over generic sampling techniques~\cite{DBLP:journals/corr/cfg,ramesh2022hierarchical}.
Our approach is inspired by classifier-free diffusion guidance~\cite{DBLP:journals/corr/cfg}, which offers a significant advantage over classifier-guided models~\cite{dhariwal2021diffusion}, balancing the diversity and fidelity of samples in conditional diffusion models. By adopting the classifier-free approach, we incorporate group interactions, global and local information to Transformer Diffusion to customize and control trajectory generation. 

\subsection{Game theory}
Game theory is proposed to examine how players or groups form coalitions to cooperate and achieve shared objectives in cooperative games. A cooperative game consists of a player set $P = \{1, 2, ..., n\}$ and a characteristic function \(\phi(\cdot)\), which maps each subset $Set_b \subseteq P$ of players to a score value, indicating the payoff when players in coalition $Set_b$ work together in the game. Mainstream methods including Shapley value~\cite{shapley:book1952,WINTER20022025} and Banzhaf value~\cite{Banzhaf_1965_5380,lehrer1988axiomatization} are widely-used characteristic function calculating the benefits of coalitions in a game. For two players $i$ and $j$ within a player set $P$, their individual rewards, \(\phi(\{i\})\) and \(\phi(\{j\})\), might be low, but if they form a coalition, the combined reward, \(\phi(\{i, j\})\), could be significantly higher. This increase occurs because the players in the coalition can interact and collaborate, potentially generating additional benefits through their cooperation.

Banzhaf Interaction~\cite{grabisch1999axiomatic,jin2023video} is a widely-used approach for valuing the benefits of coalitions in a game. Specifically, given a coalition $\{i,j\} \subseteq \mathcal{N}$, the Banzhaf Interaction $\mathcal{I}([\{i,j\}])$ for the player $[\{i,j\}]$ is defined as the following:
\begin{equation}
\begin{aligned}
\mathcal{I}([\{i,j\}])\!=\!\sum_{\!\mathcal{C} \subseteq \mathcal{N} \setminus \{i,j\} }p(\mathcal{C})[\phi(\mathcal{C}\cup \{[\{i,j\}]\})+\phi(\mathcal{C})\\
-\phi(\mathcal{C}\cup\{i\})-\phi(\mathcal{C}\cup\{j\})],
\end{aligned}
\label{apendix:BI}
\end{equation}
where $p(\mathcal{C})=\frac{1}{2^{n-2}}$ is the likelihood of $\mathcal{C}$ being sampled. "$\mathcal{N} \setminus \{i,j\} $" denotes removing $\{i,j\}$ from $\mathcal{N}$. Intuitively, $\mathcal{I}(\{i,j\})$ reflects the tendency of interactions inside $\{i,j\}$. The higher value of $\mathcal{I}(\{i,j\})$ indicates that player $i$ and player $j$ cooperate closely with each other.

As for the task of trajectory prediction in this work, especially in complex sports games, agent interactions and cooperation, driven by group strategies and competition, greatly influence individual trajectories. An agent's behavior is shaped by its intentions and others' actions, like positioning, passing, and defense. So we introduce Banzhaf Interaction to model these dynamics with game theory in a semantic approach, enhancing trajectory prediction.

\section{Proposed Method}
\subsection{Problem Formulation}
\label{ssec:problemformulation}
In this work we focus on the team sports and our objective is to utilize past trajectories and group-level information (tactic labels) from previous frames to forecast future player movements and group-level tactic labels. Taking a basketball game as an example, we formally define the problem involving \(M = 2\) teams and \(N = 11\) agents (five players from each team, plus the basketball, \(N_T = 6\)) as follows:
\begin{itemize}
\item{Let \(X^{-T_{obs}+1:0}\in\mathbb{R}^{N \times T_{obs} \times 2}\), \(X^{1:T_{pred}}\in\mathbb{R}^{N \times T_{pred} \times 2}\) represent past and future trajectories of \(N\) agents over \(T_{obs}\) and \(T_{pred}\) time steps, respectively.}
\item{Let \(L^{-T_{obs}+1:0} \in \mathbb{R}^{M \times T_{obs} \times 1}\), \(L^{1:T_{pred}} \in \mathbb{R}^{M \times T_{pred} \times 1}\) denote past and future group-level tactic labels of \(M\) teams over \(T_{obs}\) and \(T_{pred}\) time steps.}
\end{itemize}

\noindent Our goal is to develop a predictive model \( \mathcal{M} \) such that:
\[ \mathcal{M}: (X^{-T_{obs}+1:0}, L^{-T_{obs}+1:0}) \rightarrow ({X}^{1:T_{pred}}, L^{1:T_{pred}}). \]
This concise formulation encapsulates the dual prediction tasks, harnessing historical data to forecast future player movements and tactical strategies, thereby providing a comprehensive approach to game analysis and strategy development in team sports.

\textbf{Trajectory Prediction.}
Human motion trajectory prediction aims to forecast the future paths of multiple agents based on their historical movement data. Given a scenario with N agents, each agent \( i \) has historical trajectory data \( X_i = \{ x_i^t \}_{t=-T_{obs}+1}^0 \in \mathbb{R}^{T_{obs} \times 2} \), where \( i \in [1, 2, \dots, N] \) represents the identifier of the observed agents, and \( x_i^t \in \mathbb{R}^2 \) represents the position of the \( i \)-th agent at time \( t \). Our method aims to predict the future trajectories of these agents, denoted as \( Y_i = \{ y_i^t \}_{t=1}^{T_{pred}} \in \mathbb{R}^{T_{pred} \times 2}, i \in [1, 2, \dots, N] \).

\textbf{Tactic Label Prediction.}
In addition to trajectory prediction, we aim to forecast the future group-level tactic labels \( L_j^{1:T_{pred}} \) for \(j_{th}\) team with \(X^{-T_{obs}+1:0}\) and \(L_j^{-T_{obs}+1:0}\) as input, where \( j \in [1, \dots, M] \) represents the identifier of different teams in the sport game scenario.

\begin{figure*}
    \centering
    \includegraphics[width=0.9\linewidth]{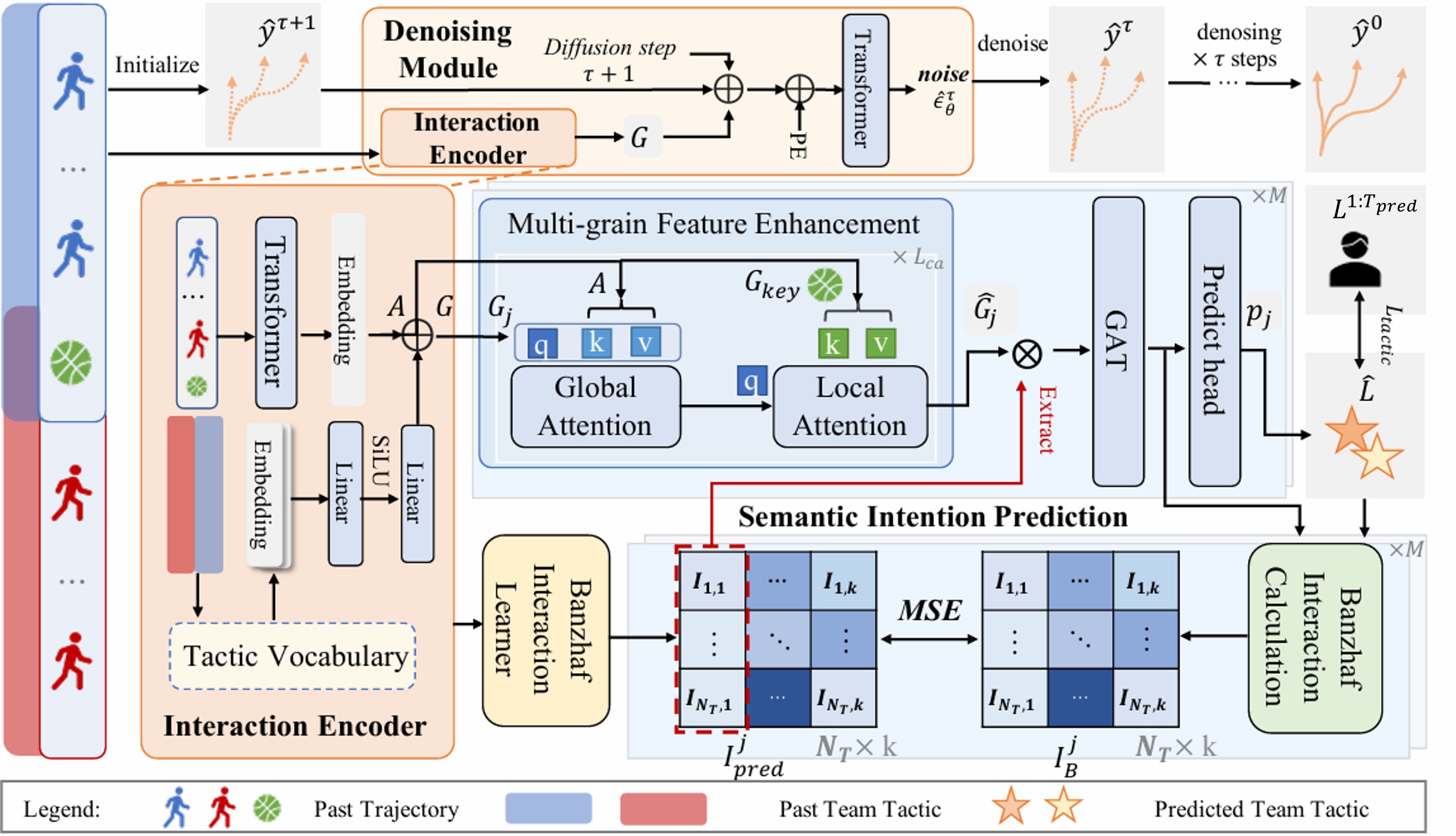}
    \caption{The overview of our proposed method. It consists of two main parts: (1) denoising module for diffusion-based trajectory prediction and (2) semantic intention prediction module for team-level tactics. Specifically, the interaction encoder processes observed trajectories and group-level tactics to generate agent tokens that serve as conditions for diffusion model to predict future trajectories. The multi-grained feature enhancement module captures global and local information to enhance agent tokens. Banzhaf Interaction Learner predicts the similarity between agents and potential Top-k tactics, which can be viewed as semantic intentions, while Banzhaf Interaction Calculation computes ground truth for supervision.
    Finally, we fuse the enhanced agent tokens with the semantic intentions and feed this information into the prediction head to obtain tactic predictions. The blue background denotes team-level predictions.}
    \label{fig:overview}
\end{figure*}

The overview of our proposed framework is shown in Figure~\ref{fig:overview}. Our framework adopts in an end-to-end manner, designed to fulfill the objectives of a two-fold prediction task, including trajectory prediction and tactic prediction. The latter one serves as an auxiliary task that enriches the primary trajectory forecasts by incorporating semantic intentions from game perspective. Firstly, we employ an interaction encoder to process observed trajectories and corresponding group-level tactics, using a classifier-free guidance diffusion model to guide trajectory generation. Additionally, we introduce a multi-grained feature enhancement module that captures both global and local information through attention mechanisms to refine agent representations. Then we model the task as a cooperative game, where agents and the tactics adopted by each group are treated as players. Furthermore, we calculate Banzhaf Interactions to represent cooperation trends within coalitions of players. To facilitate the learning of semantic intention similarities, we leverage knowledge from the Banzhaf Interaction Calculation as supervision for the Banzhaf Interaction Learner. Finally, we fuse the enhanced agent tokens with the learned semantic intention assignments, and input to the prediction head to obtain group interaction probabilities.

\subsection{Interaction Encoder}\label{sec:encoder}
Firstly, we introduce the Interaction Encoder to process observed data, including past trajectories and the given tactic labels, to generate agent tokens that represent the encoded group interactions for two-fold prediction tasks in subsequent sections. In our method, we introduce a group-level tactic vocabulary serving as a guidance for diffusion model, which includes embeddings for a comprehensive array of tactics. This is designed to facilitate nuanced understanding and application of team strategies within the predictive model. 

Initially, we employ a Transformer architecture to process the historical trajectories of all agents, denoted by \(X=\{ x_i \}_{i=1}^N \in \mathbb{R}^{N \times 2}\), to generate trajectory embeddings \(A = \{ a_i \}_{i=1}^N \in \mathbb{R}^{N \times D_A} \), encapsulating the dynamic spatial interactions among agents and their movement patterns over time. To integrate tactical considerations directly into the prediction process, we utilize the observed tactic labels for each team, represented by \(l^{-T_{obs}+1:0} \in\mathbb{R}^{M \times 1}\), and these labels allow us to extract the corresponding tactic embeddings from our learnable tactic vocabulary. Given an input tactic label \( l_j \) and an embedding matrix \( W \in \mathbb{R}^{V \times D_C} \), where \( V \) is the tactic vocabulary size (number of unique tactic indices) and \( D_C \) is the embedding dimension, the embedding operation can be expressed as:
\begin{equation}
\begin{aligned}
\Phi_{\text{tactic}}({l_j^{-T_{obs}+1:0}}) = W[l_j^{-T_{obs}+1:0}] = c_j \in \mathbb{R}^{D_C},
\end{aligned}
\end{equation}
where $\Phi_{\text{tactic}}(\cdot)$ is a embedding layer that maps the tactic label \(l_j\) to the tactic embedding \(c_j\), \( W[l_j] \) denotes the \( l_j \)-th row of the embedding matrix \( W \), \( \mathbf{c}_j \) is the resulting tactic embedding vector for tactic label \( l_j \), which has dimension as \( D_C \). Then we can obtain the expanded tactic embeddings \(C_{\text{e}}\) at scene-level according to agent-team correspondence:
\begin{equation}
\begin{aligned}
C_{\text{e}} = \left[ c_{1,m(1)}, \dots, c_{N,m(N)}\right],
\end{aligned}
\end{equation}
where each agent \( i \) is assigned the tactic embedding \( c_{i,m(i)} \), with \( m(i) \) mapping agent \( i \) to their corresponding team \( j \). We then formulate a comprehensive condition vector \(G = \left[ g_1; g_2; \dots; g_N \right] \quad \in \mathbb{R}^{N \times D_g} \), where \(D_g\) is the feature dimension, and the condition of the $i$-th agent \(g_i\) is formulated as:
\begin{equation}
\begin{aligned}
g_i = \left[ a_i ; c_{i,m(i)} \right],
\end{aligned}
\end{equation}
where \([;]\) indicates the concatenation, \( a_i \) denotes trajectory embedding for agent \(i\) and \( c_{i,m(i)} \) denotes the tactic embedding for agent \( i \).
This vector effectively integrates individual movements with team-level strategic intents, capturing the intricate interplay between individual actions and team tactics in sports scenarios. This integration also provides a solid foundation for our model to generate context-aware trajectory forecasts, enhancing its ability to anticipate and adapt to dynamic sports environments. Thus, we define the function of \textbf{Interaction Encoder} as \(G = f_\phi(X, L^{-T_{obs}+1:0})\).

\subsection{Denoising Module}
To predict trajectories, we employ a denoising module that generates future trajectories from noisy trajectories conditioned on agent embeddings from the interaction encoder, which are derived from observed trajectories and tactic labels.

We introduce observed tactic labels as additional guidance signal by adopting the Interaction Module \(f_\phi(\cdot)\). Inspired by classifier-free guidance diffusion~\cite{DBLP:journals/corr/cfg}, the reverse process of diffusion model becomes \(p_\theta(y^{t-1}|y^t,G)\), where \(G\) denotes the agent embeddings generated by the Interaction Encoder in Sec.~\ref{sec:encoder}, enhanced with team tactic information, and we interpret the output of diffusion models as the score function, the DDPM sampling procedure can be guided to sample \(y\) with high \(p(y|G)\) following specific tactic styles by:
\begin{equation}
\begin{aligned}
f_{\epsilon}(y, G) &= f_{\epsilon}(y, A) + s_g \cdot \nabla_y \log p(y|G) \\
&\propto f_{\epsilon}(y, A) + s_g \cdot (f_{\epsilon}(y, G) - f_{\epsilon}(y, A)),
\end{aligned}
\end{equation}
where \(f_{\epsilon}(\cdot)\) denotes the noise estimation module of the diffusion model, the hyperparameter \(s_g\) indicates the scale of the guidance with additional tactic information besides observed trajectory embedding. Evaluating the diffusion model with only observed trajectories as condition is done by randomly dropping out tactic information during training and replacing it with a raw observed trajectory embedding \(A\).

Then we can formulate the estimated noise and the denoised trajectory at \(\tau\)th denoising step as:
\begin{equation}
\begin{aligned}
\centering
\hat{\epsilon}^{\tau}_{\theta} & = f_{\epsilon}(\hat{y}^{\tau+1}, G, \tau + 1),  \\
\centering
\hat{y}^{\tau} &= \frac{1}{\sqrt{\alpha_{\tau}}} \left( \hat{y}^{\tau+1} - \frac{1 - \alpha_{\tau}}{\sqrt{1 - \bar{\alpha}_{\tau}}} \hat{\epsilon}^{\tau}_{\theta} \right) + \sqrt{1 - \alpha_{\tau}} \mathbf{z}, 
\end{aligned}
\label{eq:denosing}
\end{equation}
where \(\alpha_{\tau} \) and \(\bar{\alpha}_{\tau} = \prod_{i=1}^{\tau} \alpha_i \) are parameters in the diffusion process, \(\hat{y}^{\tau+1}\) is the trajectories directly at denoising step \(\tau\) provided by the initializer~\cite{mao2023leapfrog} to accelerate the inference and \(\mathbf{z} \sim \mathcal{N}(\mathbf{z}; 0, \mathbf{I}) \) is a noise. We denote the prediction result after \(\tau+1\) denoising steps as \(\hat{Y}_s\).

Afterwards, we optimize the denosing module following~\cite{gu2022stochastic} through diffusion generation by performing mean square error (MSE) loss between the output and a noise variable in standard Gaussian distribution for the current iteration \(\tau\) as in Eq.~(\ref{eq:denosing}), formulated as the following:
\begin{equation}
\mathcal{L}_{noise}=\left\|\boldsymbol{\mathbf{z}}-f_\epsilon\left(\mathbf{y}^{\tau+1}, \tau+1, f_\theta{\left(X, L^{-T_{obs}+1:0}\right)}\right)\right\|_2
,\end{equation}
where \( \epsilon \) and \( \phi \) are parameters of the diffusion model and interaction encoder respectively, and $\mathbf{z} \sim \mathbf{\left(0,I\right)}$. 

To optimize the trajectory prediction in two-fold prediction framework, we formulate trajectory prediction loss as:
\begin{equation}
\begin{aligned}
\mathcal{L}_{\text{dist}} &= \min_{s} \left\| Y - \hat{Y}_s \right\|_2, \\
\mathcal{L}_{\text{unc}} &=  \frac{\sum_s \left\| Y - \hat{Y}_s \right\|_2}{\sigma^2_{\theta} S} + \log \sigma^2_{\theta},
\end{aligned}
\end{equation}
where \(\mathcal{L}_{\text{dist}}\) constrains the minimum distance in \(S=20\) predictions using a variety loss~\cite{gupta2018socialgan} as formulated and explained in Sec.~\ref{metrics}, and \(\mathcal{L}_{\text{unc}}\) regularizes the variance \(\sigma_{\theta}\) in the initializer~\cite{mao2023leapfrog} through an uncertainty loss, balancing the diversity and accuracy of trajectory predictions while preventing them from being overly dispersed or consistent.

\subsection{Multi-Grained Feature Enhancement}

We introduce a Multi-grained Feature Enhancement Module to enhance group-level agent embeddings by aggregating global and local information, which uses global scene and local key object data (e.g., ball in sports) to enrich agent embeddings for group-level prediction, as shown in Figure~\ref{fig:overview}. The module employs both global and local attention mechanisms, using scene-level agent tokens \(G=\{g_i\}_{i=1}^N\) generated by the Interaction Encoder. Each player has an associated team index \(j\) and an agent index \(i\). In sports games (e.g., basketball), with N agents and M teams, we represent group interaction tokens for team \(j\) as \(G_j = \left[ g_{j,1}, \ldots, g_{j, N_T-1}, g_{\text{ball}} \right] \in \mathbb{R}^{N_T \times D_g)}\), where \( g_{j,i_T} \) represents the interaction feature for the \( i_T \)-th player in team \( j \), \( g_{\text{ball}} \) is the basketball feature, \(N_T-1\) is the number of agents per team, \( D_g \) is the dimension of group interaction tokens.

\textbf{Global attention mechanism} aggregates context and interactions to obtain team-level tokens. We generate query (Q), key (K), and value (V) vectors from agent tokens \(G_j\) as \(Q, K, V = {G_j}\mathbf{W_q}, {G}\mathbf{W_k}, {G}\mathbf{W_v}\). Then we can obtain the weighted sum as global interactions feature:
\begin{equation}
\begin{aligned}
\centering
   G_{j}^{'} &=
   \text{Softmax} \left( \frac{Q{K^\top}}{{\sqrt{D_k}}} \right) V,
\end{aligned}
\end{equation}
where \(\sqrt{D_k}\) is for the numerical stability of the Softmax, and \(G_{j}^{'}\) is global interactions.

\textbf{Local attention mechanism} enhances features using key object information, with the ball in basketball influencing player positioning and tactics. Similarly, we substitute the query, key and value as follows: \(Q, K, V = {G_{j}^{'}}\mathbf{W_q}, {G_{ball}}\mathbf{W_k}, {G_{ball}}\mathbf{W_v}\). Then other processing is as similar as the Global Attention Mechanism.

Finally, we can obtain the global and local fusion interactions \(\Hat{G_{j}}\) by using the global interactions \(G_{j}^{'}\) as query for the local interactions. More details of the attention mechanism are in the supplementary material.

\subsection{Semantic Intention Prediction Module} 
Interactions and cooperation among agents are crucial for determining their trajectories. An agent’s behavior is influenced by its objectives, the actions of other agents, and group strategies, such as positioning, passing, or defending in response to specific team tactics. Agents may also adapt the certain tactic based on the strategies of opposing teams. To model these complex interactions and dynamic changes in semantic intentions among agents and teams, we employ Banzhaf Interaction~\cite{grabisch1999axiomatic,jin2023video} according to the game theory, enabling more accurate multi-agent trajectory prediction. This module mainly contains three parts, as shown in Figure~\ref{fig:overview}: Tactic Prediction heads, Banzhaf Interaction Learner, and Banzhaf Interaction Calculation. To this end, we introduce an auxiliary task of tactic prediction, modeling it as a cooperative game. In this game, agents in the scene denoted as \(A = \{a_i\}_{i=1}^N\) and the potential tactics adopted by the \(j\)-th team, \(\hat{L_j} = \{\hat{l}_j\}_{j=1}^{k}\), are treated as players, where \(k\) represents the top-\(k\) potential tactics that team \(j\) will adopt.

\subsubsection{Banzhaf Interaction Learner} 

As the calculation of the exact Banzhaf Interaction is an NP-hard problem~\cite{jin2023video}, in order to speed up the computation of Banzhaf Interaction, we use Banzhaf Interaction Learner to learn the mapping from a set of agent embeddings to the similarity matrix between agents and potential tactics~\cite{grabisch1999axiomatic}. 
We adopt Banzhaf Interaction Learner \( f (\cdot) \)  to model and learn the complex interactions between agents in a multi-agent system. It leverages a combination of multi-layer perceptrons (MLPs) and self-attention mechanisms to effectively capture both local and global dependencies within the system.

Specifically, we use \(\gamma_{ctx}(\cdot)\) to encode agents' trajectory embedding to obtain context embedding. Then we utilize a self-attention layer \(\text{SA}(\cdot)\) to capture global interactions among multi agents in the scene. Then to obtain the predicted Banzhaf Interaction value, we use \(\gamma_{o}(\cdot)\) to project the high-dimensional feature vector into a set of real number. The output predicted Banzhaf Interaction value \(I_{\text{pred}}\) and overall mapping function \(f(\cdot)\) can be formulated as follows:
\begin{equation}
I_{\text{pred}} = f(A) =\gamma_{o}\left(\text{SA}(\gamma_{ctx}(A))\right)  \in \mathbb{R}^{N \times k },
\end{equation}
where we implemented \(\gamma_{ctx}(\cdot)\) and
\(\gamma_{o}(\cdot)\) using MLPs, to incorporate information for extracting social interaction features in the scene for subsequent analyse.
Then we can obtain the predicted Banzhaf Interaction \(I^j_{\text{pred}} \in \mathbb{R}^{N_T \times K }\) for \(j\)-th team, where \(N_T\) denotes the number of agent in a team plus a ball.

To regulate predicted \(I^j_{pred}\) with precise and reasonable semantic information, we use \(I^j_B \in \mathbb{R}^{N_T \times k}\), which is the output of \textbf{Banzhaf Calculation Module}, as supervision to optimize the Banzhaf Learner model, which represents the importance of each agent in the group for achieving a specific potential tactic with the highest likelihood. The detailed calculation process of \(I^j_B\) is described in Sec.~\ref{banzhaf_cal}. The training objective of \(f(\cdot)\) is formulated as follows: 
\begin{equation}
\begin{aligned}
\mathcal{L}^j_{bi} &= \left\| I^j_{\text{pred}} - I^j_{B} \right\|_2, \quad
\mathcal{L}_{bi} &= \sum_{j=1}^M \mathcal{L}^j_{bi},
\end{aligned}
\end{equation}
where it calculates the L2 norm of the difference between \(I^j_{\text{B}}\) and \(I^j_{\text{pred}}\).

\subsubsection{Tactic Prediction Head}\label{sec:tactic_pred_head}
We use the enhanced feature \( \hat{G_j} \) of team \(j\), derived from the multi-grained feature enhancement module, along with the predicted Banzhaf Interaction for each team as input. These are then processed by the tactic prediction head to generate the prediction probabilities and the Top-k predicted tactic labels for each team.

Specifically, we firstly retrieve \(I^j_{\text{pred}}\), the predicted Banzhaf Interaction between agents and Top-k potential tactic output from the Banzhaf Interaction Learner.

Then we fuse the enhanced feature \( \hat{G_j} \) with \( I^j_{\text{pred}} \) to obtain weighted agent tokens \( w_j \).
\begin{equation}
\begin{aligned}
w_j = Fusion(\hat{G_j}, I^j_{\text{pred}}),
\end{aligned}
\end{equation}
where \(Fusion(\cdot)\) represents performing a dot product.

Then we feed \( w_j \) to the the prediction head consisting of a Graph Attention Network (GAT) and MLPs (MLP), denoted as \( h_j(\cdot) \), to finally obtain tactic prediction probability \( p_j \):
\begin{equation}
\begin{aligned}
\centering
p_j &= \text{softmax} (h_j (w_j) ) \\
    &= \text{softmax} (\text{MLP} (\text{GAT}(w_j) ) \in \mathbb{R}^{V \times 1},
\end{aligned}
\end{equation}
where \(V\) represents the total number of tactics.
Then we can extract the predicted Top-k potential tactic labels \( \hat{l_j} \)and embeddings \( \hat{c_j} \):
\begin{equation} 
\begin{aligned}
\centering
\hat{l_j} &= \arg\max_{k}(p_j) \in \mathbb{R}^{k \times 1}, \\
\hat{c_j} &= \Phi_{tactic} ( \hat{l_j}) \in \mathbb{R}^{k \times D_C},
\end{aligned}
\end{equation} 
where \( \arg\max_{k} \) means extracting the Top-k values and labels based on the softmax probabilities \( p_j \), and \( \hat{l_j} \) represents the predicted Top-k tactic labels, while \( \hat{c_j} \) represents their corresponding embeddings.

We use the human-annotated tactic labels for each team in the next 20 frames as supervision, and use the focal loss~\cite{ross2017focal} for tactic prediction due to the data distribution imbalance problem~\cite{zhang2021knowledge}.
We convert future tactic labels into one-hot label and denote prediction logits $p$ as the model's estimated probability for the class where the label $y=1$. Here we define \(p_j^{tactic}\) for \(j\)-th team:
\begin{equation}
  p_j^{tactic} =
\begin{cases}
p_j^{tactic}, & \text{if}~y=1 \\
1-p_j^{tactic}, & \text{otherwise.}
\end{cases}
\label{eq:tactic_pred}
\end{equation}

Hence we formulate the tactic prediction classification focal loss for each team as follows:
\begin{equation}
\label{eq:team_focal}
    \mathcal{L}_{\text{tactic}}^j = \alpha^j\left(1-p_j^{tactic}\right)^{\gamma^j} \log\left( p_j^{tactic} \right),
\end{equation}
where $\alpha^j$ is the normalized inverse frequency of tactics, and $\gamma^j$ is a hyper-parameter. Similarly, we can express \( \mathcal{L}_{\text{tactic}} \) as a focal loss for predictive classification. The total prediction loss can be formulated as follows:
\begin{equation}
\label{eq:total_focal}
    \mathcal{L}_{\text{tactic}} = \sum_{j=1}^{M}
{\mathcal{L}_{\text{tactic}}^j}.
\end{equation}

\subsubsection{Banzhaf Interaction Calculation}\label{banzhaf_cal}
This part takes the similarity matrix of agent tokens for each group and the predicted Top-k tactic embeddings \(\hat{c_j}\) retrieved from Sec.~\ref{sec:tactic_pred_head} as inputs, which can be viewed as the interaction logits \(S\), capturing agent-tactic associations. The output can be seen as the ground-truth of Banzhaf Interaction, denoted as \(I_{\text{B}}\), where \(\mathcal{S}\) encapsulates comparative features across team tactics and their impact on the game. \(I_{\text{B}}\) quantifies each agent's influence within the team, enabling a detailed analysis of tactic interdependencies in team sports or cooperative scenarios.

Specifically, the interaction similarity \(S\) is initialized and preprocessed based on the agent and tactic masks and weights. For each pair of agent \(a\) and tactic \(t\), subsets of agent and tactic tokens are randomly selected and masked, simulating the effects of different token combinations on the interaction. The agent mask and the tactic mask are denoted as \({mask_a}\) and \({mask_t}\), respectively. The interaction logits are then computed with masking and subjected to softmax normalization. To measure how each agent influences tactic selection, we introduce the agent-to-tactic interaction logits \(S_{a2t}\), computed as follows:
\begin{equation}
\begin{aligned}
S_{a2t} =  \text{softmax}(S_{a, t} \times {mask_a}[a]),
\end{aligned}
\end{equation}
where \(a\) is the index of agent tokens, \(t\) is the index of tactic tokens, \(S\) represents the raw interaction logits, and \({mask_a}\) is the agent mask (indicating valid agent tokens).

On the other hand, to measure how each tactic influences agent behavior, such as constraining or driving agent actions, we compute the tactic-to-agent interaction logits \(S_{t2a}\) as follows:
\begin{equation}
\begin{aligned}
S_{t2a} = \text{softmax}(S_{a, t} \times {mask_t}[t]),
\end{aligned}
\end{equation}
where \({mask_t}\) is the tactic mask.

Hence, the Banzhaf Interaction for team \(j\) is computed as follows, which quantifies the importance of each agent in adopting a specific tactic within the multi-agent system:
\begin{equation}
\begin{aligned}
I^j_B = (S_{a2t} + S_{t2a}) / {2},
\end{aligned}
\end{equation}
where \(S_{a2t}\) and \(S_{t2a}\) represent the weighted logits from agent to tactic and tactic to agent, respectively. Their average yields the Banzhaf Interaction under the current masking configuration.

By calculating both influences, the model captures bidirectional interactions between agents and tactics. Averaging these measures, the Banzhaf Interaction quantifies mutual influence, providing a clearer understanding of how agents and tactics collaborate to optimize decision-making in multi-agent systems.

\subsection{Training Objective}

Based on the before-mentioned parts, the final objective function for training the end-to-end two-fold prediction framework is expressed as:

\begin{equation}
\begin{aligned}
\mathcal{L} & =\mathcal{L}_{noise} + \mathcal{L}_{\text {distance}}+\eta\mathcal{L}_{\text {unc}}+\alpha(\mathcal{L}_{\text {tactic}}+\beta\mathcal{L}_{\text {bi}}), 
\end{aligned}
\end{equation}
where $\eta$ is the hyperparameter controlling uncertainty estimation, while $\alpha$ and $\beta$ are hyperparameters that govern the influence of tactic prediction in the overall trajectory prediction task and the influence of the game prediction in the tactic prediction task, respectively.

\section{NBA Tactic Dataset}
In this section, we introduce the expanded version of the NBA SportVU dataset, including manual annotations of the strategies employed by each team during different periods of each game, aimed at enhancing group interactions.

\subsection{Data Preparation}

To incorporate detailed information about players and teams, we resample the raw SportVU data logs\footnote{https://github.com/linouk23/NBA-Player-Movements} from the 2015-2016 NBA season to obtain trajectory sequences containing the IDs of players and teams. In total, we extract 90,618 trajectory sequences for 30 frames at 5 Hz. These sequences are visualized as videos to observe the team tactics involved, which aids in data expansion.

We extend the data consisting of the following steps: First, we surveyed various NBA videos to categorize and identify popular and representative team tactics. We then select 16 widely-used types of team tactics (7 offensive and 9 defensive) for annotation in the dataset, which are commonly observed in NBA games. Offensive tactics include \emph{Pick-and-Roll}, \emph{Ball Movement}, \emph{Fast Break}, and \emph{Single}, while defensive tactics include five types of \emph{zone defense}, \emph{Man-to-Man defense}, \emph{defensive rebound}, \emph{defensive transition}, and \emph{scramble defense}. 

Using the visualized scene-level trajectories, we train 25 human annotators on expert knowledge to identify these tactics and establish a standardized annotation protocol. Finally, we manually verify the annotations to ensure their accuracy and consistency.

\subsection{Data Annotation}
We construct the extended NBA SportVU dataset, annotated with human-labeled tactics for the previous 10 frames and the following 20 frames at 5 Hz. Annotators select one tactic for each team and time period from 16 representative team tactics. For example, "zone defense" refers to a more structured and fixed formation compared to the other 8 defensive tactics, while "man-to-man defense" involves strong motion trends and more similar trajectories between offenders and their corresponding defenders. 
The dataset consists of trajectories for 11 agents per scene: five players from the offensive team, five players from the defensive team, and one basketball. Annotators' responses, which denote the tactics selected for each team during specific time intervals, are stored in a JSON file. Each entry in the file includes the scene ID, which preserves the order of the trajectory data stored in the associated .npy file, the selected tactics for each team, and the corresponding time period. In total, the dataset contains 8,238 entries, with each entry representing four selected tactics per scene (two teams). 

\begin{figure}
    \centering
    \includegraphics[width=0.9\linewidth]{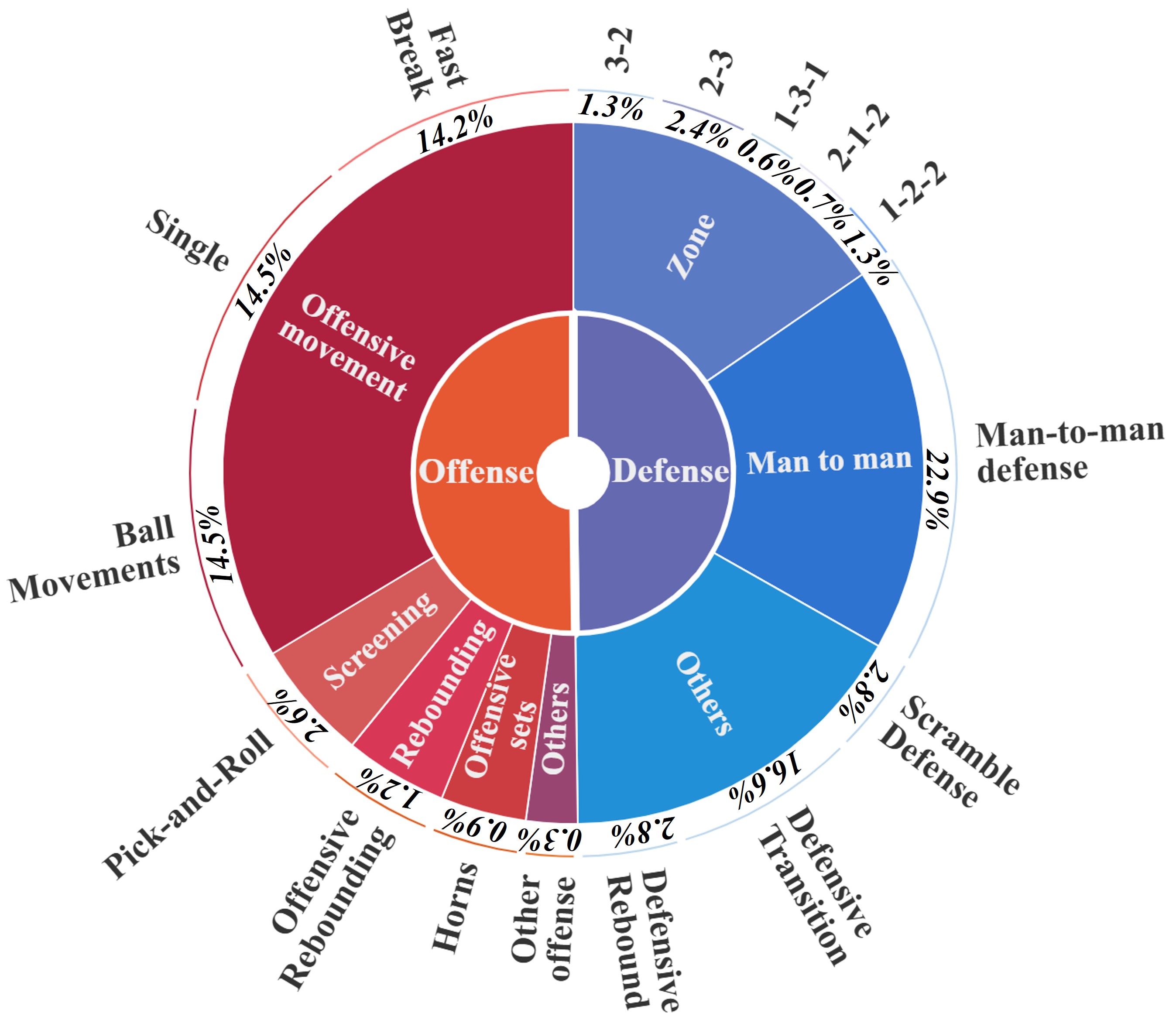}
    \vspace{-3mm}
    \caption{Illustration of the team-level tactic annotation distribution of NBA SportVU dataset.}
    \label{fig:tactic_distribution}
    \vspace{-3mm}
\end{figure}

We present team-level annotation statistics for our extended NBA SportVU dataset as follows in the Fig.~\ref{fig:tactic_distribution}. All tactics can be categorized into two primary classes: offensive and defensive. 
Each class is further subdivided into specific types by studying manuals of world association of basketball coaches\footnote{https://wabc-chn.fiba.com} and observing the team movement patterns associated with different strategies in basketball games. Offensive tactics are classified to common offensive movement, screening-based, rebounding-based, common offensive sets and others, while defensive tactics are categorized into zone defenses, man to man defense and others. More details of dataset statistics including scene-level team-wise tactic adoption statistics, and tactic transition statistics are shown in the supplementary material.

\begin{figure}
    \centering
    \includegraphics[width=0.9\linewidth]{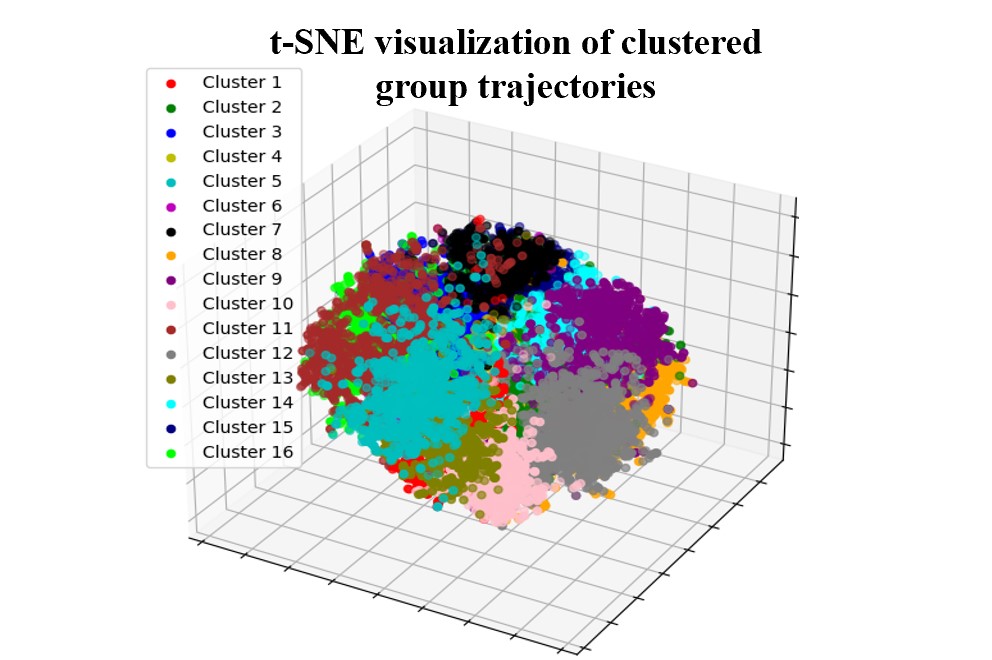}
    \vspace{-3mm}
    \caption{3D t-SNE visualized results of utilizing clustering to generate tactic pseudo-labels on NBA SportVU dataset.} 
    \label{fig:nba_pseudo}
    \vspace{-3mm}
\end{figure}

Furthermore, due to limited human resources, we introduce group-level pseudo labels as an alternative for validating trajectory prediction performance on the TeamTrack-soccer and TeamTrack-basketball datasets~\cite{DBLP:journals/corr/teamtrack}. We generate these pseudo labels using an unsupervised K-means clustering method. To reduce manual effort, we modeled our clustering approach after tactic annotation methods, grouping trajectories into 16 distinct movement patterns based on shared motion trends, which streamlined the annotation and analysis process. The 3D t-SNE visualization on the NBA SportVU dataset, shown in Figure~\ref{fig:nba_pseudo}, along with the experimental results evaluating generalizability in the supplementary material, validate the feasibility of using this clustering method to produce pseudo labels for trajectory prediction.

\section{Experiment}
\subsection{Datasets}

\textbf{NBA SportVU dataset.} Following previous literature~\cite{mao2023leapfrog}, we train our model to observe 2 seconds of past trajectory and predict 4 seconds in the future for all agents in the scene (respectively 10 and 20 frames). We split 70\% of our extended dataset for training, and the remain 30\% data for testing.

\noindent\textbf{TeamTrack dataset.} We use the TeamTrack dataset~\cite{DBLP:journals/corr/teamtrack} including basketball and soccer games to validate our method. Following~\cite{DBLP:journals/corr/teamtrack}, we utilized observed trajectories for 3.2 seconds and predicting the trajectories for the next 4.8 seconds (respectively 96 and 144 frames) in experiments.

\subsection{Metrics}
\label{metrics}

\subsubsection{Trajectory prediction}
To evaluate the performance of trajectory prediction, we use Average Displacement Error (ADE) and Final Displacement Error (FDE) as validation metric: 
\emph{Average Displacement Error (ADE)} measures the euclidean distance between prediction and the ground truth averaged by the prediction length \( T_{pred} \). 
\emph{Final Displacement Error (FDE)} calculates the L2 error between the prediction and the ground truth at the last time-step \( T_{pred} \).
Note that, we adopt a widely-used best-of-20 strategy~\cite{gupta2018socialgan,gu2022stochastic,mao2023leapfrog} during evaluation, where multiple diverse predictions are generated in 20 modes, and the minimum ADE and FDE are computed between the predictions and the ground truth. 

\subsubsection{Tactic prediction}
For the tactic prediction task, we use Top-k accuracy to evaluate the performance.
Specifically, we use the Top-1, Top-2, Top-3 and Top-5 accuracy as the evaluation metric. More details please refer to the supplementary material.

\subsection{Implementation details}
We implement our method with PyTorch~\cite{DBLP:conf/nips/PaszkeGMLBCKLGA19} on a single NVIDIA RTX 3090 GPU. We train our model with the Adam optimizer.
For both datasets, the number of training epochs is 100, and we set the diffusion step to 100. 
We set \(s_g\) in classifier free guidance diffusion to 0.1. We train the denoising module for 100 epochs with an initial learning rate 1e-3 and decay to half every 16 epochs. We train the tactic prediction model for 100 epochs with an initial learning rate 1e-3 and decay to half every 16 epochs, and the two-fold prediction framework for 30 epochs with an initial learning rate of 2e-3 and decay by 0.9 every 32 epochs. 
We set the game factor to 0.001 and the $\gamma$ in Eq.~(\ref{eq:total_focal}) to 4.0. The tactic prediction rate is set at 1000. More implementation details are in the supplementary material.

\subsection{Compared Methods} 
\textbf{Trajectory prediction.}\label{sec:trajectory_pred_baseline} For the NBA SportVU dataset, we compare our method with state-of-the-art methods as the following:  1) \textbf{S-LSTM}~\cite{alahi2016social} adopts a grid based pooling mechanism to capture the social information of surrounding agents. 2) \textbf{STGAT}~\cite{huang2019stgat} encodes spatial-temporal relationship with LSTM and Graph Attention Networks, and then use a LSTM to decode future trajectory. 3) \textbf{GroupNet}~\cite{xu2022groupnet} utilizes hyper-graph to model the interactions, and use CVAE to generate future trajectories. 4) \textbf{MID}~\cite{gu2022stochastic} encodes the history behavior information and the social interactions as a state embedding and devise a Transformer-based diffusion model to capture the temporal dependencies of trajectories. 5) \textbf{LED}~\cite{mao2023leapfrog} adopts a trainable leapfrog initializer to accelerate inference speed by directly learn an expressive multi-modal distribution of future trajectories, which skips a large number of denoising steps. 5) \textbf{SingularTrajectory}~\cite{bae2024singulartrajectory} builds a singular space to project all types of motion patterns from each task into one embedding space, and adopts a diffusion-based predictor using a cascaded denoising process. 6) \textbf{SocialCircle}~\cite{wong2024socialcircle} builds a new angle-based trainable social interaction representation SocialCircle for trajectory prediction. For the TeamTrack-trajectory datasets, we implement the following compared method: \textbf{1) Linear Velocity} represents a simple linear model that predicts the next position based on the last observed time step, using a constant velocity model. \textbf{2) LSTM motion} represents a simple LSTM-based encoder and decoder, and an MLP for mapping hidden features to 2D positions for each agent to model their motion behaviors. \textbf{3) MID} and \textbf{4) LED} are mentioned before.

\noindent\textbf{Tactic prediction.}\label{sssec:tactic_pred_baseline}
We introduce several methods for comparison, including vanilla Transformer, and large language model (LLM) based approaches. \textbf{1) Baseline} adopts the Transformer to encode past trajectories and use a two-layer MLP for prediction. \textbf{2) Baseline*} adopts the Transformer to encode past trajectories and use one-hot to encode the tactic embedding, and use a two-layer MLP for prediction. \textbf{3) Pooling*} adopts the Transformer to encode past trajectories and uses one-hot to encode the tactic embedding, then a pooling mechanism for each team to aggregate group-level information, and then we use a two-layer MLP for prediction. \textbf{4) Llama3-8B} adopts a zero-shot LLama3-8B to inference the prediction results. \textbf{5) Llama3-8B*} adopts a finetuned version of LLama3-8B with our benchmark using LoRA~\cite{hu2021lora} to inference the prediction results. We format the trajectories and annotations of our dataset in language prompt with Alpaca format. The detailed information are shown in the supplementary material.

\subsection{Quantitative Results}

\begin{table*}[]
\centering
\caption{Quantitative evaluation results of trajectory prediction on NBA SportVU dataset. $^\dagger$ denotes that values are reproduced using the official implementation. The best results are highlighted in bold.}
\vspace{-2mm}
\begin{tabular}{@{}c|ccccccc|c@{}}
\toprule
    Method & S-LSTM & STGAT & GroupNet & MID & LED$^\dagger$ & SingularTraj$^\dagger$ & SocialCircle$^\dagger$ & Ours \\ 
    Time & CVPR2016 & CVPR2019 & CVPR2022 & CVPR2022 & CVPR2023 & CVPR2024 & CVPR2024  & Tactic-e2e\\ 

\midrule
1.0s        & 0.45/0.67 & 0.38/0.55 & 0.34/0.48 & 0.28/0.37 & 0.21/0.31 & 0.28/0.44 & 0.45/0.61         & \textbf{0.19}/\textbf{0.29} \\
2.0s        & 0.88/1.53 & 0.73/1.18 & 0.62/0.95 & 0.51/0.72 & 0.42/0.63 & 0.61/1.00 & 0.67/0.90              & \textbf{0.39}/\textbf{0.59} \\
3.0s        & 1.33/2.38 & 1.07/1.74 & 0.87/1.31 & 0.71/0.98 & 0.65/0.92 & 0.96/1.47 &0.99/1.25            & \textbf{0.61}/\textbf{0.86} \\
4.0s & 1.79/3.16 & 1.41/2.22 & 1.13/1.69 & 0.96/1.27 & 0.89/1.24 & 1.31/1.98 & 1.18/1.46    & \textbf{0.84}/\textbf{1.17} \\ 
\bottomrule
\end{tabular}
\label{tab:nba}
\vspace{-3mm}
\end{table*}

\begin{table*}[htbp] 
\centering 
\caption{Quantitative evaluation results of trajectory prediction on TeamTrack-Soccer and TeamTrack-Basketball datasets. \\The best results are highlighted in bold.} 
\vspace{-2mm}
\begin{tabular}{c|cccc|cccc} 
\toprule 
Method & \multicolumn{4}{c|}{TeamTrack-Soccer} & \multicolumn{4}{c}{TeamTrack-Basketball} \\
\midrule 
\(T_{pred}\)& 1.2s & 2.4s & 3.6s & 4.8s & 1.2s & 2.4s & 3.6s & 4.8s \\
\midrule 
Linear & 0.93 / 1.89 & 1.97 / 4.11 & 3.10 / 6.62 & 4.32 / 9.27 & 1.01 / 2.03 & 2.08 / 4.18 & 3.15 / 6.38 & 4.26 / 8.75 \\
LSTM & 2.32 / 3.99 & 4.02 / 7.18 & 5.50 / 9.49 & 6.72 / 11.22 & 1.37 / 2.30 & 2.27 / 3.96 & 3.08 / 5.34 & 3.80 / 6.51 \\
MID & 0.33 / 0.61 & 0.89 / 2.00 & 1.65 / 3.57 & 2.44 / 4.81 & 0.56 / 1.08 & 1.29 / 2.57 & 2.07 / 3.76 & 2.74 / 4.28 \\
LED & 0.27 / 0.52 & 0.68 / 1.30 & 1.19 / 2.16 & 1.75 / 3.01 & 0.49 / 0.76 & 1.00 / 1.61 & 1.50 / 2.33 & 2.01 / 3.99 \\
\midrule 
Ours & \textbf{0.23} / \textbf{0.45} & \textbf{0.58} / \textbf{1.10} & \textbf{1.00} / \textbf{1.89} & \textbf{1.48} / \textbf{2.72} & \textbf{0.42} / \textbf{0.72} & \textbf{0.88} / \textbf{1.44} & \textbf{1.31} / \textbf{2.05} & \textbf{1.72} / \textbf{2.78} \\
\bottomrule 
\end{tabular}%
\label{tab:combined_teamtrack}%
\vspace{-3mm}
\end{table*}

\textbf{Trajectory prediction.} We report the detailed results obtained by our proposed method on NBA SportVU dataset and comparison to previous state-of-the-arts in Table~\ref{tab:nba}. We can see that our proposed method outperforms other candidates in terms of minADE/minFDE. Specifically, compared to MID~\cite{gu2022stochastic}, LED~\cite{mao2023leapfrog} and SingularTrajectory~\cite{bae2024singulartrajectory} which utilize the Transformer diffusion framework and focus solely on simple scene-level agent interaction modeling, our method outperforms all of them across all metrics. It surpasses the state-of-the-art method, LED~\cite{mao2023leapfrog} by large margins (up to 9.5\% and 6\% in minADE for the first second and total four seconds, respectively). 
It validates the importance of integrating group interactions into diffusion models to guide trajectory generation and capture dynamic semantic intentions from a game-theoretic perspective. Furthermore, we also conduct experiments on the TeamTrack-Soccer and TeamTrack-Basketball datasets, utilizing pseudo-labels for group-level guidance to predict trajectories. The corresponding quantitative evaluation results are presented in Table~\ref{tab:combined_teamtrack}. Our method outperforms the previous state-of-the-art, LED~\cite{mao2023leapfrog}, by 15.4\% and 9.6\% on the ADE and FDE metrics for the TeamTrack-Soccer dataset. Similarly, it outperforms LED~\cite{mao2023leapfrog} by 14.4\% and 30.3\% on the ADE and FDE metrics for the TeamTrack-Basketball dataset, respectively, over a total of 4.8 seconds. These results demonstrate the effectiveness of group-level guidance in our method, leading to a reduction in cumulative errors.

\textbf{Tactic prediction.} We report the evaluation results of the auxiliary tactic prediction in Table~\ref{tab:tactic_pred}. To ensure fair comparisons, we employ the same Transformer as the context encoder and the same training scheme across all methods. We compare our model against three classical baselines and two LLM-based baselines. 
It is important to note that the task is a prediction, not a classification, and we use only the observed information to predict future tactics. Our method outperforms all other baselines in each metric, improving Top-1 accuracy by 5\% over the Pooling method and 22\% over the fine-tuned Llama3-8B model. Although LLMs possess common-sense knowledge and reasoning abilities, they are not adept at comprehending trajectory sequence data or inferring group interactions effectively in complex sports games. This demonstrates the superiority of our approach in understanding dynamic group-level semantic intentions and predicting changes in group interactions from trajectory data by introducing knowledge of Banzhaf Interaction. 

\begin{table}[htbp]
  \centering
  \caption{Quantitative evaluation results of tactic prediction on NBA SportVU dataset. We introduce LLM to validate the performance of our prediction. Llama3-8B means zero-shot model, Llama3-8B* means using LoRA to finetune with our dataset.}
  \vspace{-2mm}
    \begin{tabular}{c|ccccc}
    \toprule
    Method (\%) &  Top-1  & Top-2  & Top-3  & Top-5 \\
    \midrule
    Baseline & 34.56 & 59.77 & 73.78 & 88.53 \\
    Baseline*  & 53.31 & 71.64 & 81.99 & 91.78 \\
    Pooling  & 55.22 & 75.82 & 86.05 & 94.67 \\
    
    \midrule
    Llama3-8B  & 22.39 & - & - & - \\
    Llama3-8B*  & 38.17 & - & - & - \\
    \midrule
    Ours  & \textbf{60.23} & \textbf{81.15} & \textbf{90.53} & \textbf{96.79} \\
    \bottomrule
    \end{tabular}%
  \label{tab:tactic_pred}%
  \vspace{-3mm}
\end{table}%

\begin{figure*}
    \centering
    \includegraphics[width=0.9\linewidth]{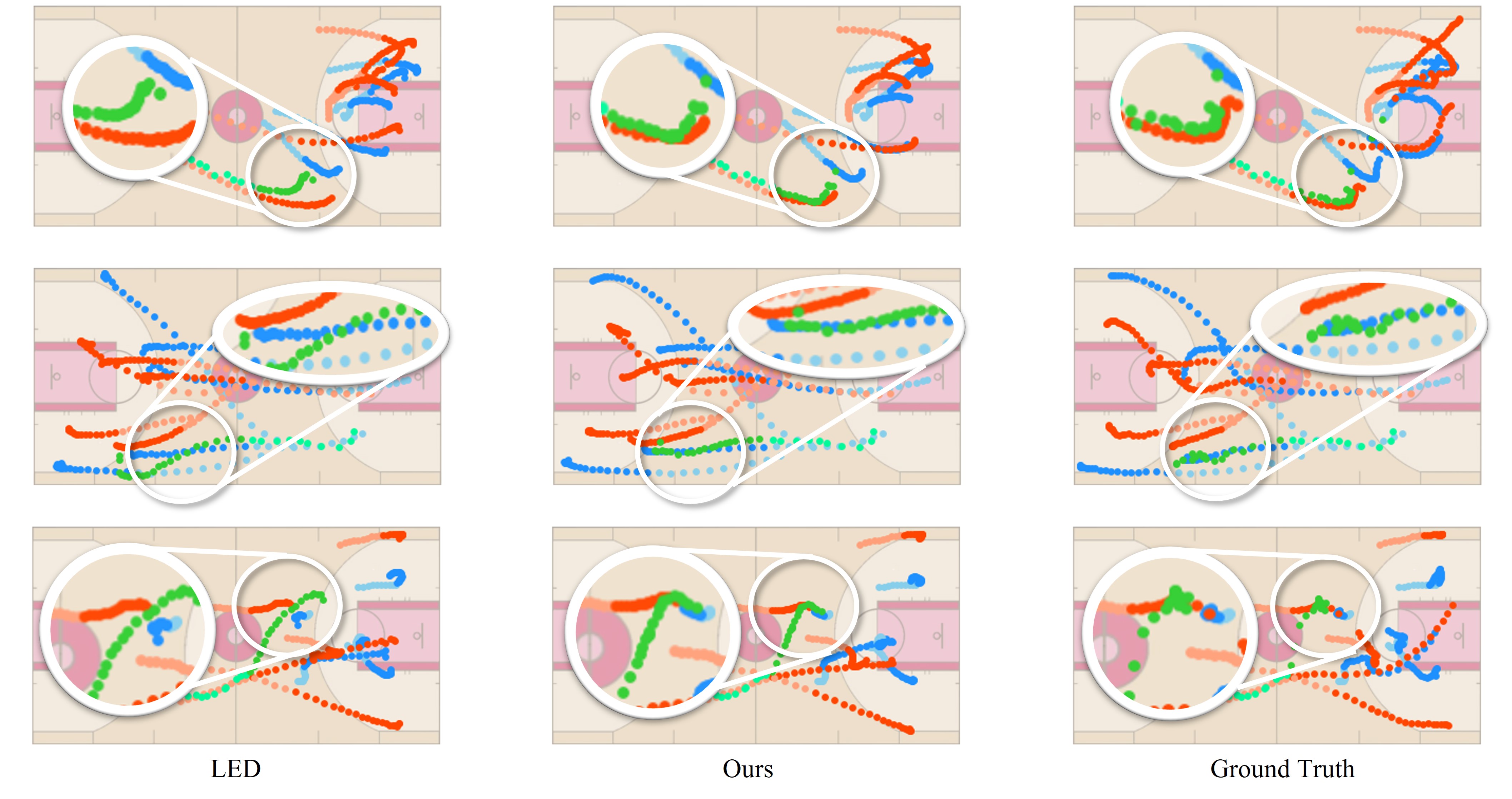}
    \vspace{-3mm}
    \caption{Illustration of visualization results of trajectory prediction on the NBA SportVU dataset, where the ground truth show in the last column, LED method~\cite{mao2023leapfrog} and our proposed approach is show in the first column and the second column, respectively. Main differences are zoom in for highlight.}
    \label{fig:traj_pred_case}
    \vspace{-2mm}
\end{figure*}

\subsection{Qualitative Results}

\begin{figure}
    \centering
    \includegraphics[width=1.0\linewidth]{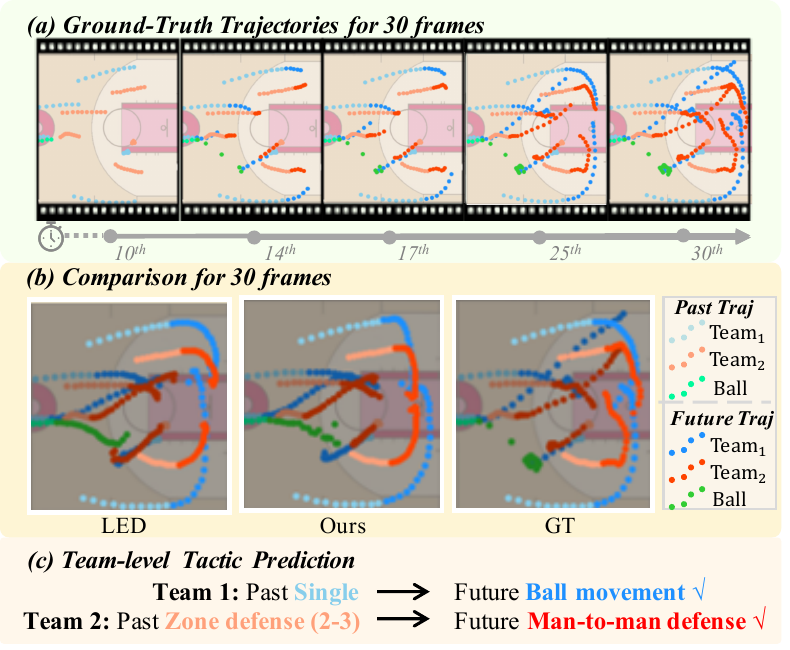}
    \vspace{-5mm}
    \caption{Qualitative results of two-fold prediction tasks on NBA SportVU dataset. The past trajectories of the first team, second team, and basketball are shown in light blue, light orange, and light green, respectively, with future trajectories in dark blue, red, and dark green.
    The first row shows the ground truth across frames, while the second row compares future trajectories with key interactions (e.g., "man-to-man defense" by Team 2 and "ball movement" by Team 1), and the bottom row displays tactic prediction results.}
    \label{fig:vis_overview_pred}
    \vspace{-3mm}
\end{figure}

We present a two-fold prediction qualitative example on the NBA SportVU dataset in Figure~\ref{fig:vis_overview_pred}. 
The comparison of future trajectories is shown in the second row, with specific trajectories and group interactions highlighted for clarity. 
We also show the tactic prediction results for each team of our method.
The first team (depicted in blue) employed the "single" tactic for the past 10 frames. Our method predicts correctly that the tactic for the next 20 frames will be "ball movement" with a notably higher probability than the others.
Meanwhile, the second team starts with the "Zone defense (2-3)" tactic but switches to "Man-to-man defense" for the next 20 frames, which our model successfully predicts. This tactic change is largely due to the first team's tactic threatening their defense and the ball handler's proximity to the basket, prompting a strategic shift to a more flexible and compact defense. As a result, the trajectories of the defensive players mirror those of the offensive players, as shown in the trajectory prediction comparison. In contrast, the prediction from LED~\cite{mao2023leapfrog} displays an unclear "man-to-man defense" motion pattern, with the corresponding players positioned too far apart. This suggests that LED~\cite{mao2023leapfrog} lacks the ability to model group interactions and understand semantic intentions. However, our model effectively captures and predicts this pattern, using it to guide future trajectory predictions and yielding more realistic, interactive results that closely align with the ground truth. 

Furthermore, we present additional visual comparisons of generated trajectories in Fig.~\ref{fig:traj_pred_case}. 
Our proposed method outperforms the LED method in trajectory prediction, delivering superior accuracy and flexibility. It effectively captures complex player interactions, rapid directional changes, and tactical coordination. Specifically, our method offers more accurate predictions of player passes in the three scenarios presented in the figure. In contrast, the LED method generates more linear predictions, with significant discrepancies between player and ball trajectories, struggling with rapid movements and failing to capture tactical understanding, often resulting in deviations when agents change positions or directions abruptly. More qualitative results are shown in the supplementary material.

\begin{figure*}
    \centering
    \includegraphics[width=0.8\linewidth]{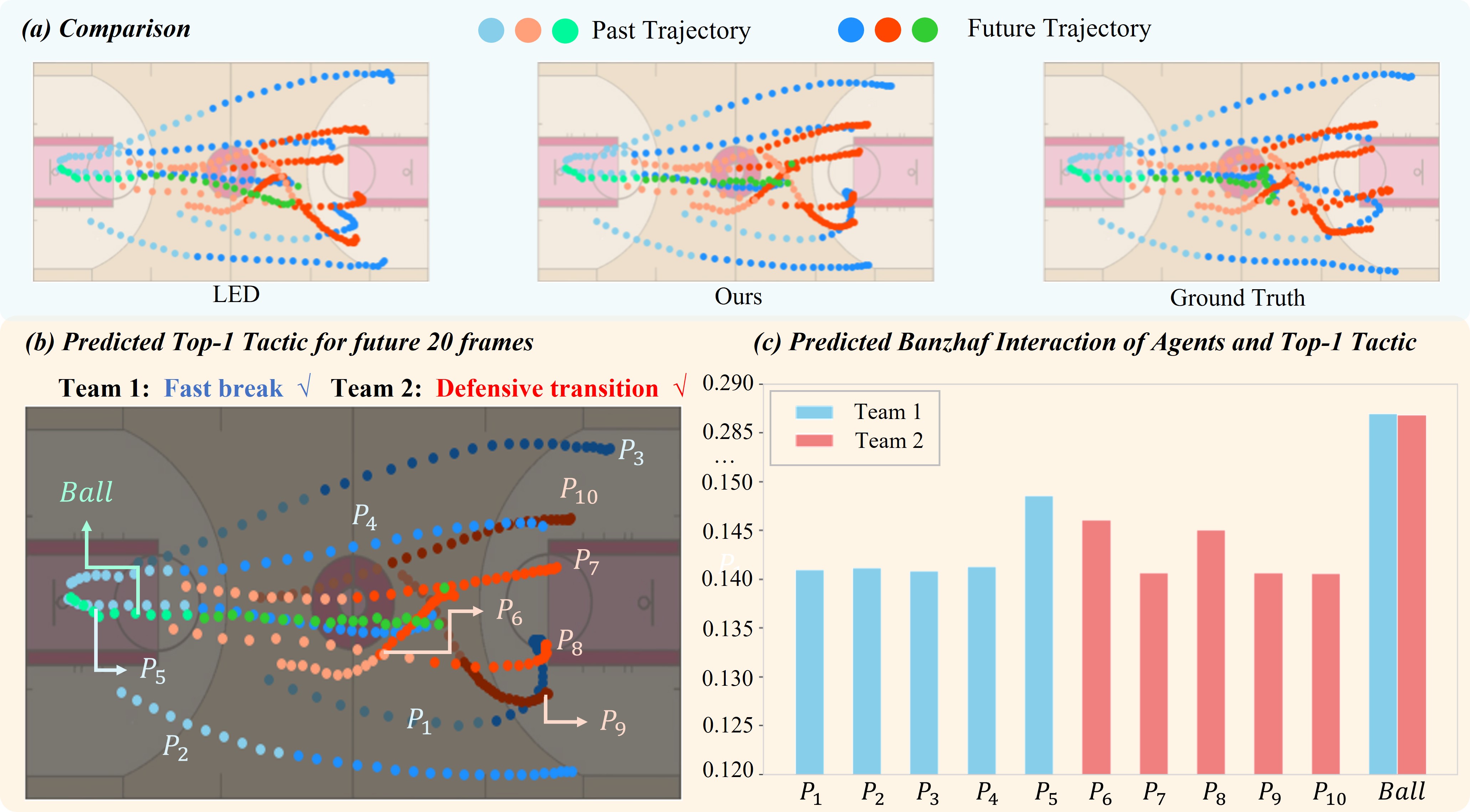}
    \vspace{-2mm}
    \caption{The visualization results of the predicted Banzhaf Interaction on the NBA SportVU dataset. This figure shows the ground truth (in the last column) alongside results from various methods for direct comparison, including LED~\cite{mao2023leapfrog} (in the first column) and our proposed method (in the second column). The predicted Banzhaf Interaction for each agent on each team is also provided, with agents who have relative higher Banzhaf Interactions for the predicted top-1 tactic highlighted in the figure. The detailed values are shown in the bar chart.
    }
    \label{fig:bi_pred_case}
    \vspace{-3mm}
\end{figure*}

\subsection{Ablation Studies}
We conduct the ablation studies to validate the effectiveness of each design in our method, including Denoising Module and Semantic Intention Prediction Module, and the results are shown in each section. The experiments are conducted on the NBA SportVU dataset and we keep all the other settings the same.

\subsubsection{Denoising Module}

To validate the importance of a diffusion-based framework, we implement two baselines as variants: an MLP-based method and a query-based method. The MLP-based method encodes and decodes trajectories using MLPs, while the query-based method employs learnable queries with an attention mechanism to generate agent tokens for trajectory prediction. Both methods use the same interaction encoder as ours. We report the experimental results in Table~\ref{tab:ablation_framework}, and we can clearly observe that our proposed diffusion-based method outperforms alternative approaches, primarily due to its ability to handle temporal dynamics of group interactions and semantic changes effectively in complex environments like sports games, while the query-based model struggles with long-term interactions due to static queries and attention mechanisms.

\begin{table}[htbp]
  \centering
  \caption{Comparison of trajectory prediction frameworks using minADE/minFDE metrics, including MLP-based, query-based, and diffusion-based methods. The best results are highlighted in bold.}
  \vspace{-3mm}
    \begin{tabular}{c|cc|c} 
    \toprule
    \(T_{pred}\) & MLPs & Query-based & Diffusion-based \\

    \midrule
    1.0s  & 1.80/2.97 &0.37/0.56 & \textbf{0.19/0.29} \\
    2.0s  & 3.15/5.44 &0.67/1.00 & \textbf{0.39/0.59}\\ 
    3.0s  & 4.31/7.34 &0.96/1.34 & \textbf{0.61/0.86}\\ 
    4.0s & 5.28/8.71 &1.24/1.63 & \textbf{0.84/1.17}\\
    \bottomrule
    \end{tabular}%
  \label{tab:ablation_framework}%
  \vspace{-3mm}
\end{table}%

\begin{table}[htbp]
  \centering
  \caption{Ablation study of tactic prediction on the NBA SportVU dataset. "BI" refers to the Banzhaf Interaction module, while "Type" indicates the type of Banzhaf Interaction Learner we design, which includes both team-level and scene-level models.}
  \vspace{-3mm}
    \begin{tabular}{cc|cccc}
    \toprule
    BI  & Type     & Top-1  & Top-2  & Top-3  & Top-5 \\
    \midrule
    $\times$ & -  & 58.44 & 78.44 & 88.11 & 94.89 \\ 
    \checkmark & Team & 60.09 & 80.99 & 90.43 & 96.75 \\
    \midrule
    \checkmark & Scene &  \textbf{60.23} & \textbf{81.15} & \textbf{90.53} & \textbf{96.79} \\
    \bottomrule
    \end{tabular}%
  \label{tab:ablation_game}%
  \vspace{-3mm}
\end{table}%

\subsubsection{Semantic Intention Prediction Module}
We explore the role of game theory in our method, comparing two Banzhaf Interaction Learner variants: (1) the \textbf{Scene}-level, which uses tokens from the entire scene to predict interactions, and (2) the \textbf{Team}-level, which processes tokens for each team separately. 
Experimental results show in Table~\ref{tab:ablation_game}, and we can see that the Banzhaf Interaction module improves Top-1 to Top-5 accuracy by 2\%. In addition, the scene-level method outperforms the team-level by capturing broader agent interactions across the scene, leading to better predictions of agents' intentions and behaviors.

Furthermore, we present a case to valid our proposed Banzhaf Interaction. The visualization of trajectory prediction and Banzhaf Interaction on the NBA SportVU dataset is shown in Fig.~\ref{fig:bi_pred_case}.
From the figure, over 30 frames, the first team (blue) executes a fast-break tactic toward the second team (red), which is transitioning to defense. The Banzhaf Interaction is calculated for the five players and the ball. The ball has the highest interaction value (about 0.285) for both teams. For the first team, Player 5, holding the ball, has the highest Banzhaf Interaction, highlighting the key role. For the second team, Player 6 and Player 8, positioned near the ball handler, show the highest values, indicating their critical role in defending. This visualization demonstrates the effectiveness and interpretability of our Banzhaf Interaction-based semantic intention modeling.

\section{Conclusion and Future Works}
In this work, we presented a novel diffusion-based trajectory prediction method, utilizing group interactions and semantic intention among agents. We also build a benchmark by annotating tactics on NBA SportVU dataset for two-fold prediction tasks. Extensive experiments demonstrates the effectiveness and superiority of our method. In the future, we aim to extend our proposed approach to more tasks such as multi-object tracking and sports video captioning.

{\small
    \bibliographystyle{IEEEtran}
    \bibliography{sample}
}

%







\end{document}